\documentclass[10pt,twocolumn,letterpaper]{article}

\usepackage{iccv}
\usepackage{authblk}
\usepackage{times}
\usepackage{epsfig}
\usepackage{graphicx}
\usepackage{amsmath}
\usepackage{amssymb}
\usepackage{enumitem}
\usepackage{multirow}
\usepackage{booktabs}

\usepackage[T1]{fontenc}
\usepackage[utf8]{inputenc}
\usepackage[english]{babel}
\usepackage[autostyle, english=american]{csquotes}
\MakeOuterQuote{"}

\usepackage[breaklinks=true,bookmarks=false]{hyperref}

\iccvfinalcopy 

\def\iccvPaperID{****} 

\ificcvfinal\fi

\begin{document}

\title{The Solution for the ICCV 2023 1st Scientific Figure Captioning Challenge}


\author{
  Dian Chao$^1$,
  Xin Song$^2$,
  Shupeng Zhong$^1$,
  Boyuan Wang$^1$,
  Xiangyu Wu$^1$,
  Chen Zhu$^3$,
  Yang Yang\thanks{Corresponding author: Yang Yang(yyang@njust.edu.cn)} $^1$
}

\affil{
  $^1$Nanjing University of Science and Technology
  $^2$Baidu,Inc
  $^3$Boss Zhipin
}

\maketitle
\ificcvfinal\thispagestyle{empty}\fi

\begin{abstract}
  In this paper, we propose a solution for improving the quality of captions generated for figures in papers. We adopt the approach of summarizing the textual content in the paper to generate image captions. Throughout our study, we encounter discrepancies in the OCR information provided in the official dataset. To rectify this, we employ the PaddleOCR toolkit to extract OCR information from all images. Moreover, we observe that certain textual content in the official paper pertains to images that are not relevant for captioning, thereby introducing noise during caption generation. To mitigate this issue, we leverage LLaMA to extract image-specific information by querying the textual content based on image mentions, effectively filtering out extraneous information. Additionally, we recognize a discrepancy between the primary use of maximum likelihood estimation during text generation and the evaluation metrics such as ROUGE employed to assess the quality of generated captions. To bridge this gap, we integrate the BRIO model framework, enabling a more coherent alignment between the generation and evaluation processes.  Our approach ranked first in the final test with a score of 4.49.
\end{abstract}

\section{Introduction}

The inclusion of figures and tables in a research paper plays a pivotal role in enhancing the readers' comprehension of complex knowledge presented within the paper. To obtain higher-quality captions, we explore various methods for generating high-quality titles for these figures, including the utilization of multimodal large models like OFA~\cite{ofa} and BLIP~\cite{blip}, which leverage both image and text information for title generation~\cite{yang2019semi,yang2019semi2}. We also attempted fine-tuning the LLaMA~\cite{llama} large model with the expectation of improving title quality by harnessing the extensive pre-training on textual data. However, our experimental results indicate that these methods were less effective than the official baseline, which relies on text summarization.

To improve the handling of longer text inputs, we utilized the PegasusX~\cite{pegasusx} abstractive model. Focusing on the importance of mentions. We separated mentions from paragraphs and concatenated text into OCR+mention+paragraph segments for summary generation. We also enhanced data quality by using PaddleOCR for OCR extraction. To address distracting information in paragraphs, we employed the Llama-2-7B model, improving model effectiveness. To control sentence quality, we integrated the BRIO~\cite{brio} model, which generates summaries and evaluates candidate outputs based on Rouge-2-normalized scores. This modification optimized performance on the Rouge-2-normalized metric.

Our main contributions can be summarized as follows:

\begin{itemize}[itemsep=0pt,parsep=0pt,topsep=0pt,partopsep=0pt,leftmargin=*]
    \item \textbf{Enhance OCR Quality: }The original dataset had inaccuracies in its OCR data, we improved the quality of OCR information by utilizing PaddleOCR. 
    \item \textbf{Refine Paragraph with LLaMA:} We addressed the challenge of multiple figures and excessively long text in paragraphs by using the Llama model to distill the text, making it more concise and relevant.
    \item \textbf{Address Exposure Bias~\cite{bias}:} We incorporated the Brio model, to address exposure bias during text generation and testing evaluation. 
\end{itemize}

\section{Related Work}

Image captioning is a cross-disciplinary problem that combines computer vision and natural language processing. Approaches to address this task can be categorized into two main types: those based on image captioning and those based on generating text associated with images.

\textbf{Image Captioning.} This approach aims to transform images into natural language captions that describe their content. Classical methods involve using Convolutional Neural Networks (CNNs) to extract image features and then employing Recurrent Neural Networks (RNNs) or models like Transformers~\cite{trans} to generate textual descriptions. For example, model such as OFA (One For All)~\cite{ofa} are representative model that focus on merging text and image to generate rich image captions. Recently, our team has been experimenting with semi-supervised methods to generate image captions, and employing noise self-perception to complete missing words~\cite{yang2022exploiting,yang2021rethinking}.

\textbf{Text Generation with Image Context.} This category of approaches not only considers the image but also leverages textual information associated with the image. These methods often make use of large-scale language models such as the GPT~\cite{gpt} series or Pegasus~\cite{pegasus} to generate natural language descriptions related to images. These models perform exceptionally well in handling complex text and image information due to their ability to capture rich associations between text and images.

In the field of text summarization, several classical models are designed to condense extensive text into concise summaries, such as Pegasus and BertSum~\cite{bertsum}. For example, Pegasus stands out as a powerful pre-trained language model that excels in text summarization tasks. It employs self-attention mechanisms and masked language modeling objectives, enabling outstanding performance in generative summarization tasks.

\begin{figure}[t]
\begin{center}
   \includegraphics[width=1.0\linewidth]{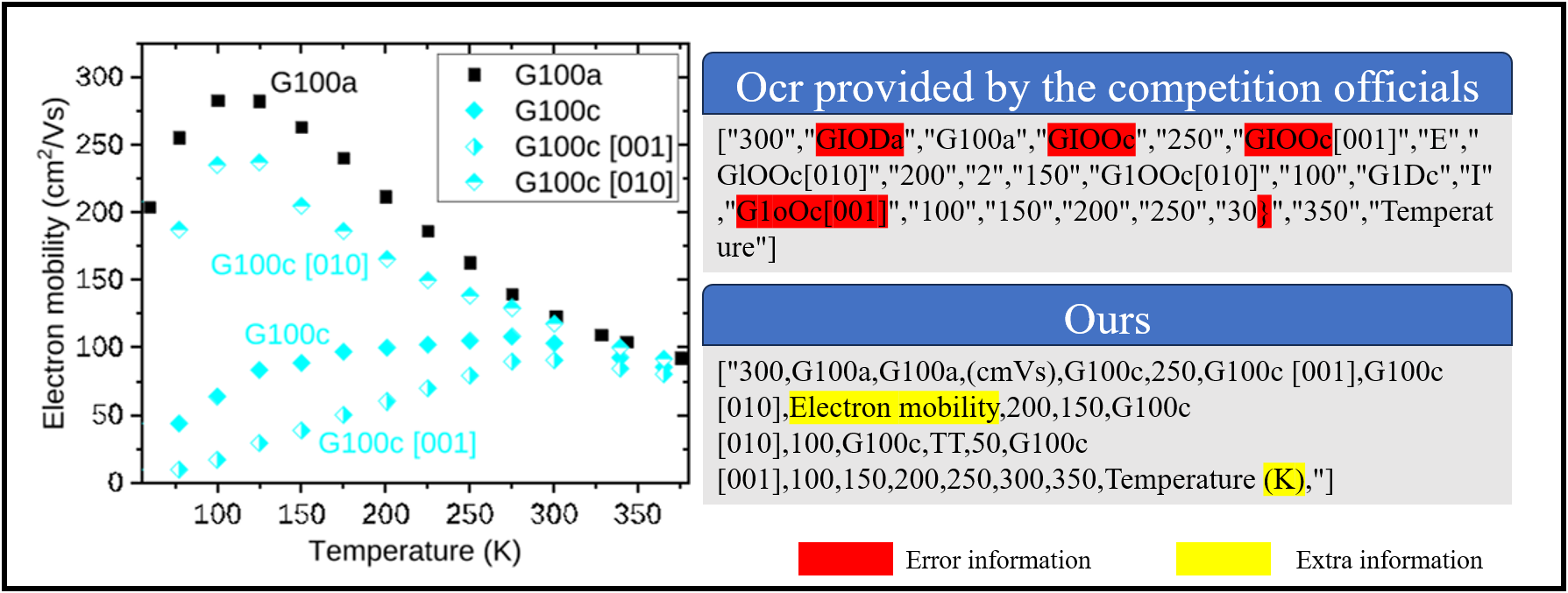}
\end{center}
   \caption{In the official dataset, OCR information contains errors, such as extracting ‘300’ as ‘30\}’ Our OCR not only corrects these errors but also extracts more valuable information from the figures.}
\label{fig:long}
\label{fig:onecol}
\end{figure}

\begin{figure}[t]
\begin{center}
   \includegraphics[width=1.0\linewidth]{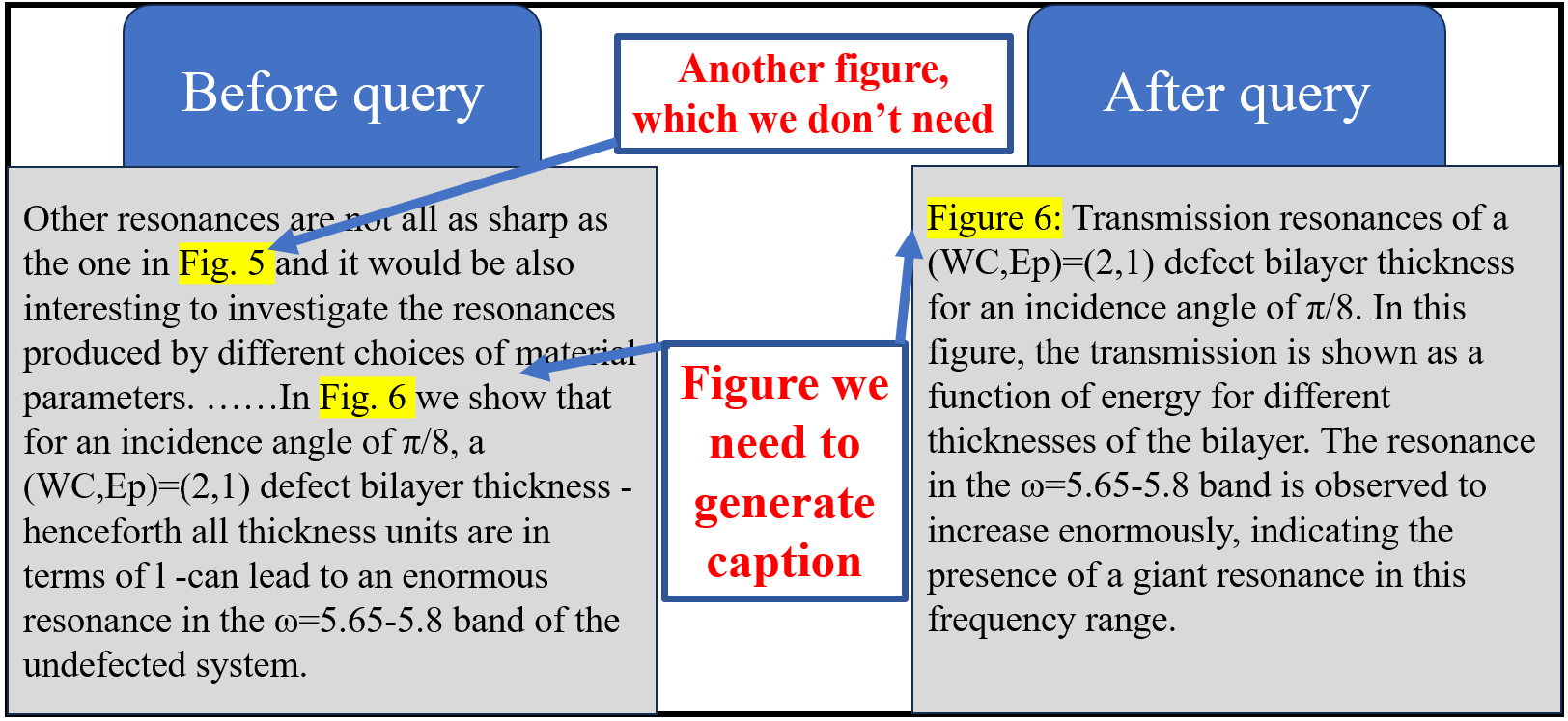}
\end{center}
   \caption{Using LLaMA to query paragraphs and refine results: Initially, the presence of information from multiple other images in the original paragraph caused interference. However, after querying the paragraph with LLaMA, only the information pertaining to the target image is retained.}
\label{fig:long}
\label{fig:onecol}
\end{figure}

\section{Method}
\subsection{Overall Architecture}
Figure 3 illustrates the overall framework of our solution. We introduce a caption generation approach that leverages high-quality image OCR information, refined paragraph information, and incorporates contrastive learning techniques. Additionally, we design a strategy to combine results from multiple diverse models, thereby achieving optimal outcomes.

\subsection{Image information extraction}
As shown in Figure 1, there are issues with information inaccuracies and important details missing in the OCR information provided by the official source. Consequently, we employed PaddleOCR to re-extract more accurate image information. Specifically, we utilized the PP-OCRv3 model to re-recognize the images in the paper, enhancing the accuracy of the OCR information.

\begin{figure*}
\begin{center}
   \includegraphics[width=1.0\linewidth]{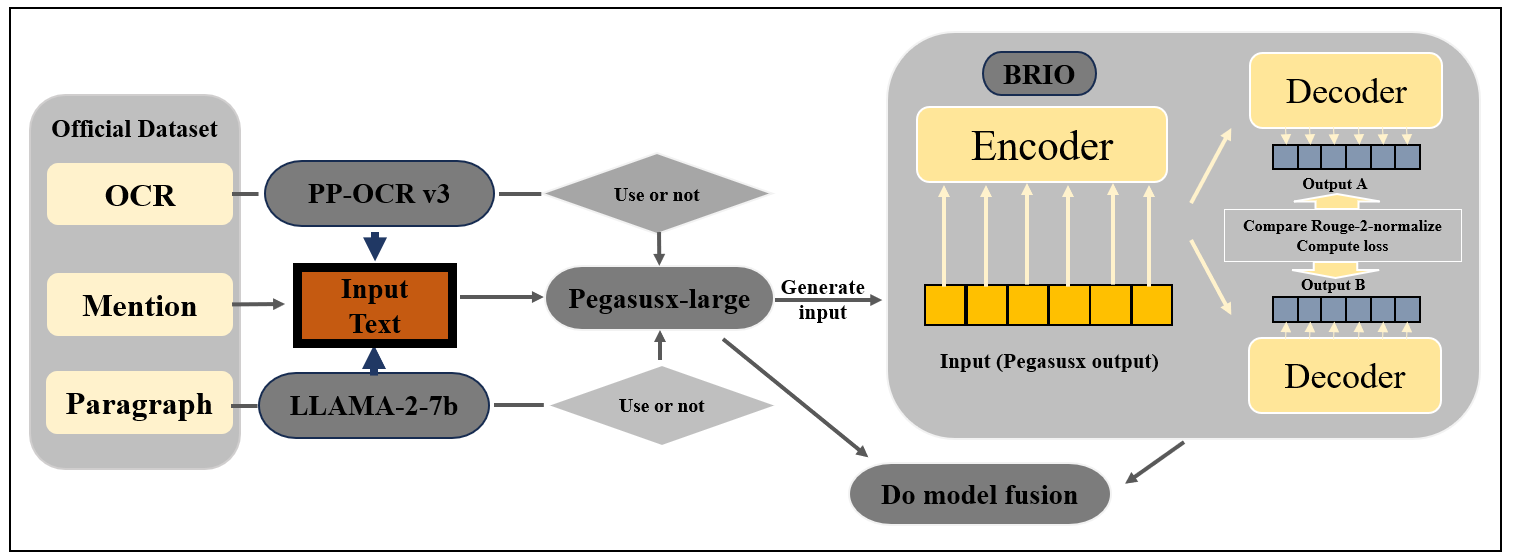}
\end{center}
   \caption{The main structure of our approch.}
\label{fig:short}
\end{figure*}
\subsection{Text information refinement}
In the course of our dataset analysis, we identified that certain data samples had paragraph segments with excessive character counts, which posed challenges for model training. Additionally, we observed that some paragraph segments contained information related to multiple images, as illustrated in Figure 2. We believed that this could introduce interference during model training. To mitigate this interference and reduce the character count in paragraph segments, we employed the state-of-the-art LLaMA-2-7b language model for summarization.

Specifically, we first determined the image numbers of interest from the 'mention' section. Subsequently, we utilized these image numbers to query the paragraph segment.

\textbf{Prompt: }The content I provide includes two sections, namely ‘paragraph’ and ‘mention’. ‘Paragraph’ and ‘mention’ are data related to figures or tables in a paper. According to the most mentioned figure in the ‘mention’ section, provide detailed information about this figure from the ‘paragraph’ section!
\subsection{BRIO Model}

Typically, conventional summarization models are trained using Maximum Likelihood Estimation (MLE), which maximizes the predicted probability of reference outputs given the golden reference sequences that precede them. However, during inference, the model must also generate outputs based on potentially erroneous preceding steps. This can compromise model performance, a phenomenon commonly referred to as exposure bias. To address this, the BRIO model is proposed.

In recent years, contrastive learning has gradually been applied to optimize generative summarization. Contrastive learning can enhance a model's performance on specific evaluation metrics by introducing richer information, such as diversity and novelty in candidate summaries. The BRIO model we employ takes a pretrained Pegasus-large model as its base model and incorporates the concept of contrastive learning during summary generation training to enhance its performance on the Rouge-2-normalized metric.

We begin by training a well-performing PegasusX model using preprocessed data. Subsequently, we use this PegasusX model to generate candidate summaries for the training data, producing four candidate summaries for each data point. This dataset serves as the training data for the BRIO model.

The goal of abstractive summarization is to create a function $g$ that takes a source document $D$ and generates an appropriate summary $S$

\begin{align}
    S \leftarrow g(D)
\end{align}

In this work we define it as the Rouge-2-normalized score of a candidate summary $S_i$ given the reference summary $S^*$. following the previous work~\cite{rank}: 

\begin{align}
    \mathcal{L}_{ctr} = \sum_{i}\sum_{j>i}max(0,f(S_j)-f(S_i)+\lambda_{ij})
\end{align}

where $S_i$ and $S_j$ are two different candidate summaries and $ROUGE(S_i, S^*) > ROUGE(S_j, S^*)$, $S^*$ is the reference summary, $\forall i, j, i < j$. $\lambda_{ij}$ is the margin multiplied by the difference in rank between the candidates, i.e., $\lambda_{ij} = (j - i) * \lambda$. $f(S_i)$ is the length-normalized estimated log-probability

\begin{align}
    f(S) = \frac{\sum_{t=1}^{l}\log(p_{g_\theta}(s_t | D, S_{<t}; \theta)}{|S|^\alpha}
\end{align}

where $\alpha$ is the length penalty hyperparameter, $\theta$ denotes the parameters of $g$ and $p_{g_\theta}$ denotes the probability distribution entailed by these parameters, $s_t$ is the next word.

Finally, we combine the contrastive and cross-entropy losses to preserve the generation ability of the pre-trained abstractive model:

\begin{align}
    \mathcal{L}_{mul} = \mathcal{L}_{xsent} + \gamma \mathcal{L}_{ctr}
\end{align}

The trained BRIO model is then used as the final model to generate answers for the test dataset.

\subsection{Model fusion}
At the end, we integrated the results from multiple models. Specifically, we performed combinations of whether to use PaddleOCR, whether to use refined paragraph information, and whether to use Brio for contrastive learning. This resulted in the training of multiple models. Subsequently, we evaluated these models on the test dataset and merged the generated results. 

The fusion method involves the following steps: for the same target caption, we have a total of eight candidate captions generated by different models. We transform this into a question: selecting the caption with the highest Rouge-2-normalized value from these eight captions. To achieve this, we calculate the Rouge-2-normalized value between each candidate caption and the other seven captions, then take the average as the score for each candidate caption. Finally, we select the candidate caption with the highest score as our answer.

The formula is as follows:

\begin{align}
    argmax_{i \in [0,N-1]} = \frac{\sum_{j \neq i}^{N}R(caption_i,caption_j)}{N}
\end{align}

Where $R(caption_i,caption_j)$ is defined as the Rouge-2-normalized score of $caption_j$ with respect to $caption_i$. We select the caption with the highest score as the final caption.

\section{Experiment}
\textbf{Dataset.} This dataset is provided by the official competition organizers and comprises approximately 400,000 data samples. Each data sample consists of an image and its corresponding OCR information. The dataset also includes mention information referring to the image as mentioned in the source paper, as well as the entire paragraph from which the mention originates. Specifically, the training set contains 333,472 data samples, while both the validation set and the public and hidden test sets consist of 47,639 samples each.

\textbf{Implementation Detail.} In our study, we trained the PegasusX-large pre-trained model. The training was conducted using an A6000 GPU, and the optimal performance was achieved at epoch 3. Subsequently, the trained PegasusX model was employed to generate candidate summaries for the Brio model, which were then used for further training. We employed two A6000 GPUs for training the Brio model. The learning rate was set to 0.00001, with a batch size of 2, and training was conducted for 50 epochs.

\textbf{Result.} Our results are presented in Table 1, where ‘Base’ represents the outcome of training with the PegasusX-large pre-trained model for three epochs. ‘+’ signifies the integration of the corresponding methods on top of the baseline. ‘combine’ denotes the amalgamation of these methods, followed by optimization using the algorithm mentioned in Section 3.4. We observe that although there is a slight decrease in Blue4, Rouge-1, and Rouge-2 scores after optimization with the Brio model, Rouge-1-normalized and Rouge-2-normalized exhibit significant improvements, affirming the effectiveness of our approach.

\begin{table}
\centering
\begin{tabular}{cccccccc}
\toprule
Method & Blue4 & R-1 & R-2 & R-1-n & R-2-n \\
\hline
Base & 0.11 & 0.46 & 0.28 & 2.18 & 3.806 \\
+Paddleocr & 0.11 & 0.44 & 0.27 & 2.18 & 3.987 \\
+llama & 0.1 & 0.43 & 0.27 & 2.21 & 4.1 \\
+brio & 0.08 & 0.41 & 0.24 & 2.21 & 4.08 \\
combine & 0.08 & 0.41 & 0.25 & 2.33 & 4.49 \\
\toprule
\end{tabular}
\caption{Optimization and results based on the Pegasusx.}
\end{table}

\section{Conclusion}
This report summarizes our solution for the  ICCV 2023 1st Scientific Figure Captioning Challenge. In this paper, we elucidate the integration of contrastive learning principles in the abstract generation process, which enhances the accuracy of OCR-based image extraction, reduces input text length, eliminates redundant information, and effectively improves the quality of summaries.

{\small
\bibliographystyle{ieee_fullname}
\bibliography{egpaper_final}
}
\end{document}